\pdfoutput=1

\documentclass[11pt]{article}

\usepackage{array}
\usepackage{multirow}
\usepackage{booktabs}
\usepackage{caption}
\usepackage{subcaption}
\usepackage{geometry}
\usepackage{url}

\usepackage[final]{acl}

\usepackage{times}
\usepackage{latexsym}

\usepackage[T1]{fontenc}

\usepackage[utf8]{inputenc}

\usepackage{microtype}

\usepackage{inconsolata}

\usepackage{graphicx}
\usepackage{hyperref}
\usepackage{fontawesome5}
\usepackage[figuresright]{rotating}
\usepackage[most]{tcolorbox}
\definecolor{BoxBackground}{RGB}{248, 248, 248} 
\definecolor{BoxFrame}{RGB}{0, 0, 0} 
\definecolor{TitleBackground}{RGB}{0, 0, 0} 
\definecolor{TitleText}{RGB}{255, 255, 255} 
\tcbset{
  academicbox/.style={
    fontupper=\footnotesize,
    boxsep=5pt,
    left=2pt,
    right=2pt,
    bottom=0.5pt,
    boxrule=0.5pt,
    colback=BoxBackground,
    colframe=BoxFrame,
    colbacktitle=TitleBackground,
    coltitle=TitleText,
    enhanced,
    attach boxed title to top left={yshift=-0.1in,xshift=0.1in},
    boxed title style={boxrule=0pt,colframe=white},
    title={#1},
  }
}
\newtcolorbox{AcademicBox}[1][]{academicbox=#1}
\usepackage[most]{tcolorbox}
\usepackage{listings}
\usepackage{tabularx}
\usepackage{adjustbox}
\usepackage{makecell}

\definecolor{mygreen}{rgb}{0,0.6,0}
\definecolor{myblue}{rgb}{0.1,0.1,0.9}

\newcommand{\highlight}[1]{%
  \tcbox[enhanced, boxrule=0pt, left=0pt, right=0pt, top=0pt, bottom=0pt, 
         colback=orange!10, colframe=orange!10, sharp corners]{\strut #1}%
}

\newcommand{\orangecell}[1]{%
  \tcbox[enhanced, boxrule=0pt, left=0pt, right=0pt, top=0pt, bottom=0pt, 
         colback=orange!10, colframe=orange!10, sharp corners, on line]{\strut #1}%
}
\newcommand{\greencell}[1]{%
  \tcbox[enhanced, boxrule=0pt, left=0pt, right=0pt, top=0pt, bottom=0pt, 
         colback=green!10, colframe=green!10, sharp corners, on line]{\strut #1}%
}

\lstdefinestyle{custompython}{
  language=Python,
  basicstyle=\ttfamily\small,
  keywordstyle=\color{mygreen}\bfseries,
  emph={openfactcheck}, emphstyle=\color{myblue}\bfseries,
  showstringspaces=false,
  breaklines=true
}

\tcbset{
  mybox/.style={
    colback=white,
    colframe=black,
    listing only,
    listing options={style=custompython},
    boxrule=0.5pt,
    arc=0mm,
    left=1mm,
    right=1mm,
    top=1mm,
    bottom=1mm,
  }
}
\definecolor{s_doc_qa_c}{HTML}{da0d68}
\definecolor{m_doc_qa_c}{HTML}{da1d23}
\definecolor{summarization_C}{HTML}{ebb40f}
\definecolor{dialogue_c}{HTML}{187a2e}
\definecolor{synthetic_c}{HTML}{0aa3b5}

\title{LOOM-Scope: a comprehensive and efficient LOng-cOntext Model evaluation framework}

\author{Zecheng Tang$^{1,2}$, Haitian Wang$^{1,2}$, Quantong Qiu$^{1,2}$, Baibei Ji$^{1,2}$ \\
\vspace{3pt}
\textbf{Ruoxi Sun}$^{1,2}$\textbf{,} \textbf{Keyan Zhou}$^{1,2}$\textbf{,}  \textbf{Juntao Li}$^{1,2}$\thanks{\; Corresponding author.}\textbf{,} \textbf{Min Zhang}$^{1}$ \\
$^{1}$Soochow University, China \\
$^{2}$Key Laboratory of Data Intelligence and Advanced Computing, Soochow University \\
\texttt{zctang@stu.suda.edu.cn}~~~\texttt{\{ljt,minzhang\}@suda.edu.cn} 
}

\begin{document}
\maketitle
\begin{abstract}
Long-context processing has become a fundamental capability for large language models~(LLMs). 
To assess model's long-context performance, numerous long-context evaluation benchmarks have been proposed. 
However, variations in evaluation settings across these benchmarks lead to inconsistent results, making it difficult to draw reliable comparisons. 
Besides, the high computational cost of long-context evaluation poses a significant barrier for the community to conduct comprehensive assessments of long-context models.
In this paper, we propose LOOM-Scope, a comprehensive and efficient framework for long-context evaluation. 
LOOM-Scope standardizes evaluation settings across diverse benchmarks, supports deployment of efficient long-context inference acceleration methods, and introduces a holistic yet light-weight benchmark suite to evaluate models comprehensively.\footnote{Homepage: \href{https://loomscope.github.io}{\faGithub\  \url{https://loomscope.github.io}}} 

\end{abstract}

\begin{figure}[t]
  \includegraphics[width=\columnwidth]{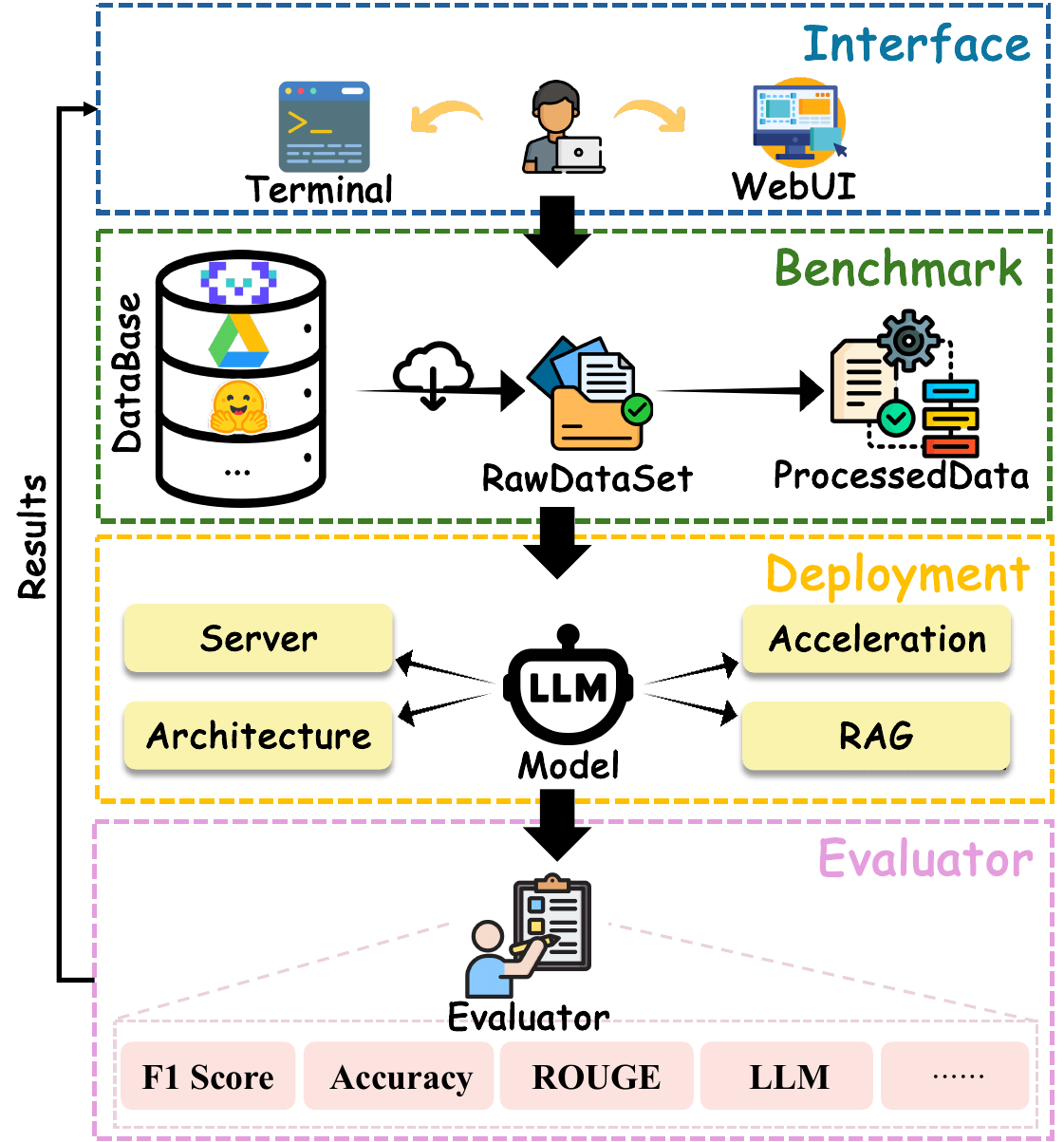}
  \caption{Workflow of LOOM-Scope framework.}
  \vspace{-5pt}
  \label{fig:loom_overview}
\end{figure}

\section{Introduction}
\label{sec:intro}
The ability to process long context is an essential capability for large language models~\citep{bertsch2024context,meta2025llama}, enabling them to address previously challenging areas like long-context reasoning~\citep{kuratov2024babilong} and unlock unbounded external knowledge through the long context input~\citep{ray2025survey}.
Alongside advancements in long-context language models~(LCLMs), recent research in the long-context processing field has increasingly focused on two directions: (1) establishing various benchmarks to evaluate model performance across diverse long-context tasks~\citep{liu2025comprehensive} and (2) improving efficiency for long-context inference~\citep{zhou2024survey}.

Yet, with the increasing number of benchmarks, inconsistent evaluation results across benchmarks create substantial obstacles in assessing and selecting appropriate LCLMs for different usage.
For example, when evaluating the model's long-context reasoning capability on different benchmarks, \textit{LongBench V2}~\citep{bai2024longbench} indicates that GLM-9B~\citep{glm2024chatglm} slightly outperforms Llama-3.1-8B~\citep{meta2024introducing}, while \textit{LongReason}~\citep{ling2025longreason} shows Llama-3.1-8B significantly surpasses GLM-9B.
Furthermore, as benchmarks increasingly incorporate longer context distributions and expand their domain coverage, the evaluation requires significant computational resources, e.g., simply evaluating Llama-3.1-8B on \textit{RULER}~\citep{hsieh2024ruler} benchmark even costs over 100 H20~(92GB) GPU hours, posing a serious challenge for the community.


To mitigate the aforementioned issues, we introduce \textbf{LOOM-Scope}, a comprehensive and efficient framework for \textbf{LO}ng-c\textbf{O}ntext \textbf{M}odel evaluation. 
As shown in Figure~\ref{fig:loom_overview}, LOOM-Scope supports two usage modes: via terminal and WebUI, and consists of three key modules: the \textsc{Benchmark} module, the \textsc{Deployment} module, and the \textsc{Evaluator} module.
Within each module, the evaluation environment is fully customized to minimize confounding factors across benchmarks, such as instruction prompts and inference hyperparameters, enabling \textbf{fair and comparable assessments} among benchmarks. 
In addition, to enhance \textbf{inference efficiency} while maintaining model performance, LOOM-Scope supports three representative long-context optimization techniques: retrieval-augmented generation (RAG)~\citep{Liu_LlamaIndex_2022}, key-value cache optimization~\citep{li2024survey}, and training-free sparse attention~\citep{zhang2025sageattention}.
Our framework is also compatible with efficient inference frameworks such as vLLM~\citep{kwon2023efficient} and SGLang~\citep{zheng2024sglang}.

Our platform supports 22 long-context benchmarks and more than 140 long-context tasks. 
To demonstrate the effectiveness of our platform, we construct a comprehensive evaluation set for experiments, \textsc{LOOMBench}, by up-sampling from 12 mainstream open-source long-context benchmarks.
With LOOM-Scope, this evaluation can be completed very efficiently: for models of 8B scale, a full capability assessment covering 6 different competencies on \textsc{LOOMBench} requires only approximately 50 H20 GPU hours or 140 RTX 3090 (24GB) GPU hours, significantly less than the computational cost demanded by most single-capability long-context benchmarks.


\begin{figure*}[t]
  \includegraphics[width=\linewidth]{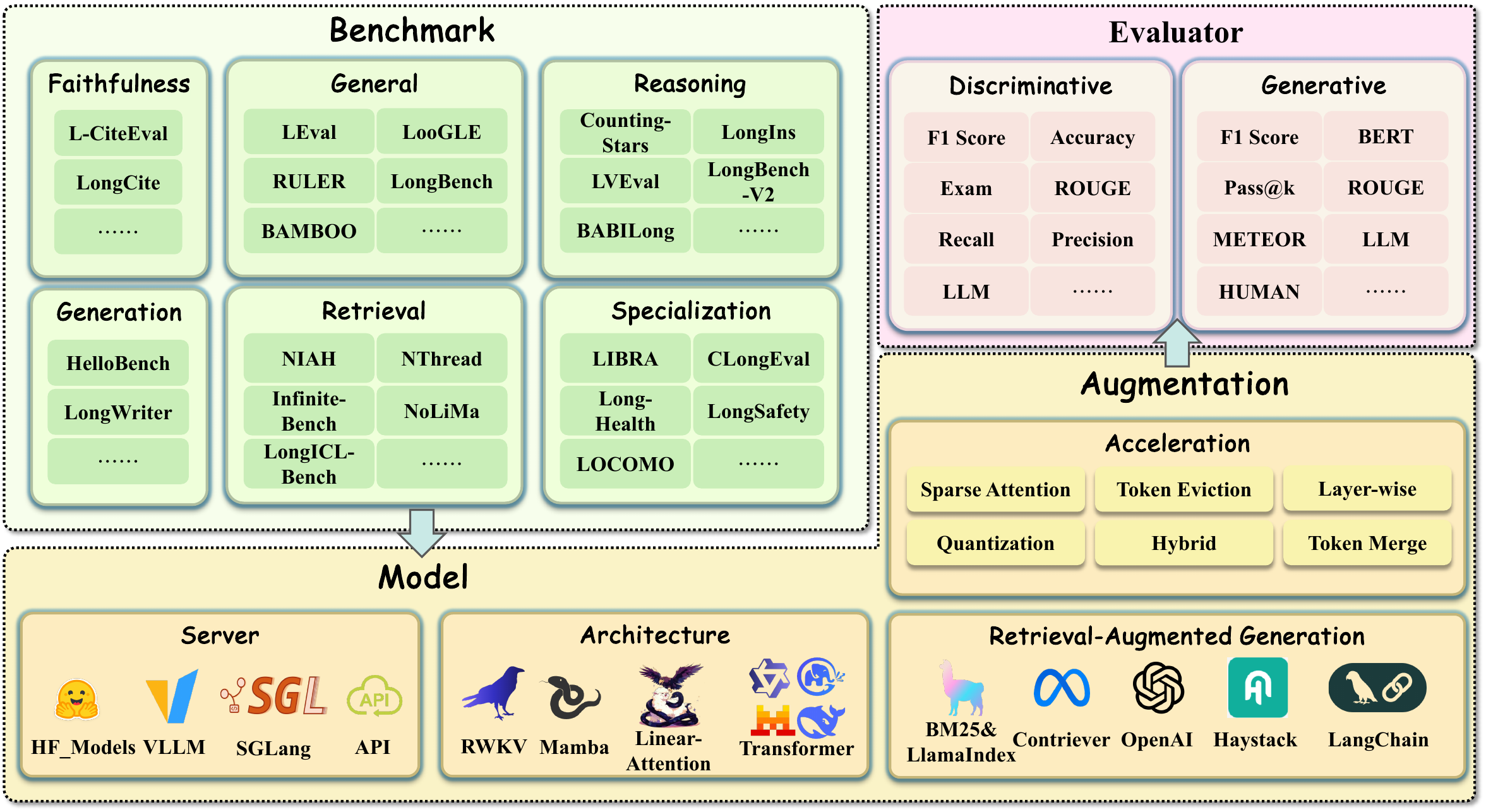}
  \vspace{-1em}
  \caption {Overview of LOOM-Scope, where the workflow of three modules can be refereed to Figure~\ref{fig:loom_overview}.}
  \label{fig:loom_scope_overview}
\end{figure*}
\section{Framework Design}
\label{sec:framework_design}
As shown in Figure~\ref{fig:loom_scope_overview}, the LOOM-Scope platform consists of three key modules, including \textsc{Benchmark} module, \textsc{Deployment} module, and \textsc{Evaluator} module.

\subsection{\textsc{Benchmark} Module}
The \textsc{Benchmark} module allows for automatic benchmark data detection, downloading, raw data pre-processing, data structure converting.
To support efficient distributed inference, the \textsc{Benchmark} module automatically allocates an appropriate number of samples to each GPU, ensuring distributed inference balance.
Currently, LOOM-Scope supports \textbf{22} widely-used long-context benchmarks, covering over \textbf{140} sub-tasks, \textbf{8k$\sim$2M} context range, and spanning \textbf{6} major LCLM capabilities: General, Faithfulness, Reasoning, Retrieval, Generation, and Specialization.
Details of supporting benchmarks are shown in Appendix~\ref{appdix:bench_overview}.
To \textbf{ensure fair evaluation} and eliminate performance discrepancies caused by prompt variations across benchmarks, \textsc{Benchmark} module allows users to define a instruction template applicable to each task.

\subsection{\textsc{Deployment} Module}
Following data processing in the \textsc{Benchmark} module, the \textsc{Deployment} module handles the deployment and inference of models.
It supports diverse model architectures~(\textsc{Model} sub-module) and advanced inference optimization techniques~(\textsc{Augmentation} sub-module), including inference servers and augmentation methods. 
Lists of models and optimization strategies are shown in Appendix~\ref{appdix:bench_overview}~(Table~\ref{tab:deployment_module}).

\paragraph{\textsc{Model} Sub-module}
This sub-module aims to support diverse server infrastructures and model architectures, including RWKV~\citep{peng2023rwkv}, Mamba~\citep{mamba}, Linear-Attention~\citep{yang2024fla}, and Transformer-based models. 
It also supports standardized deployment interfaces, including HF\_Models~\citep{wolf-etal-2020-transformers}, VLLM~\citep{kwon2023efficient}, SGLang~\citep{zheng2024sglang}, and API, ensuring flexible integration and scalability for long-context processing.


\paragraph{\textsc{Augmentation} Sub-module}
To enhance the long-context inference, the \textsc{Augmentation} sub-module supports various inference optimization techniques such as Sparse Attention~\citep{lou2024sparser}, KV-cache optimization~\citep{goel2025caote,hooper2024kvquant}, and RAG-based augmentation~\citep{leng2024long} methods. 
Specifically, Sparse Attention~\citep{laiflexprefill,xu2025xattention,jiang2024minference} accelerates inference by selectively focusing on relevant tokens, reducing computational overhead. 
KV-cache optimization~\citep{SnapKV,KIVI,GEAR} significantly reduces memory usage and enhances inference speed. 
RAG-based augmentation~\citep{bm25,Liu_LlamaIndex_2022} leverages retrieval mechanisms to enhance model external knowledge, improving performance on tasks requiring extensive context. This sub-module maintains consistency with the \textsc{Model} sub-module, while supporting custom user-defined methods for comprehensive evaluation.

\subsection{\textsc{Evaluator} Module}
After obtaining all the model predictions, the \textsc{Evaluator} module serves as a comprehensive assessment. 
It integrates both discriminative and generative evaluation metrics to provide a holistic view of the model's capabilities in long-context processing.
Discriminative metrics, including F1 Score, Accuracy, Exam, Recall, Precision, and LLM-specific evaluations, are integrated to assess the model's understanding ability.
To assess the model's generation capability, metrics such as BERT, ROUGE, Pass@k, METEOR, and human evaluations are employed to gauge the quality, coherence, and relevance of the generated text. 
The module ensures that the outputs of the \textsc{Deployment} module are rigorously tested against these diverse metrics.
Statistics of the evaluation metrics for each benchmark are shown in Appendix~\ref{appdix:bench_overview}~(Table~\ref{tab:benchmarks}).

\subsection{Workflow of LOOM-Scope}
As shown in Figure~\ref{fig:loom_overview}, the above three modules are finally integrated into the following evaluation workflow: (1) verifies the availability of and pre-process the specified benchmarks~(\textsc{Benchmark} module), (2) deploys and run LCLMs with specified sever and augmentation methods~(\textsc{Deployment} module), and (3) conducts evaluation after model prediction~(\textsc{Evaluator} module).
\begin{figure}[t]
\begin{AcademicBox}[\footnotesize LOOM-Scope Usage]
\textcolor{blue}{\textit{\textbf{Command Line}}}
\vspace{5pt}\hrule \vspace{1pt}
\begin{verbatim}
loom-scope.run \
    --model_path {model path} \
    --cfg_path {benchmark config} \
    --template {template config} \
    --device {device id} \
    --gp_num {data parallel size} \
    --server {server config} \
    --acceleration {augmentation config} \
    --eval \
    --save_tag {save name}
\end{verbatim}
\vspace{2pt}
\textcolor{blue}{\textit{\textbf{Local Web Interface}}}
\vspace{4pt}\hrule \vspace{2pt}
\begin{verbatim}
python WebUI/app.py # will open a gradio
\end{verbatim}
\end{AcademicBox}
\vspace{-1em}
\caption{Illustration of two low-code ways to start customized evaluation with LOOM-Scope.}
\label{fig:command_line}
\end{figure}

\begin{figure*}[t]
  \includegraphics[width=\textwidth]{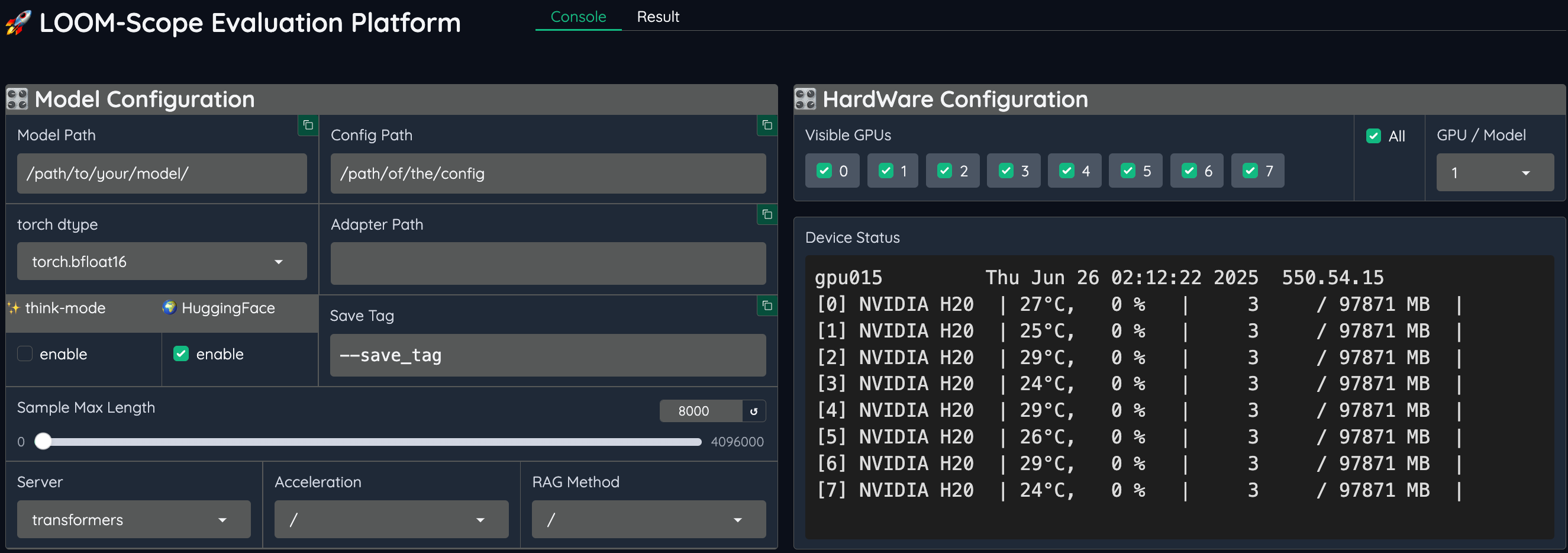}
  \caption{Partial snapshot of the local deployment WebUI of LOOM-Scope.}
  \vspace{-1em}
  \label{fig:screen_shot_webui}
\end{figure*}

\section{Usage and Experiments}
\label{sec:usage_and_exp}
LOOM-Scope is accessible via both the command line and the local WebUI interface.
For efficient evaluation of LCLM’s capabilities, LOOM-Scope also contains an integrated benchmark~\textsc{LOOMBench}, which is up-sampled and reorganized from 12 different available benchmarks, allowing for comprehensive evaluation of an 8B LCLM within 6 hours.
In Appendix~\ref{appdix:statistic_loombench}, we show the task distribution of \textsc{LOOMBench} in Figure~\ref{fig:loombench} and the statistics of each sub-task as well as the evaluation methods.

\begin{figure}[t]
    \centering
    \includegraphics[width=\linewidth]{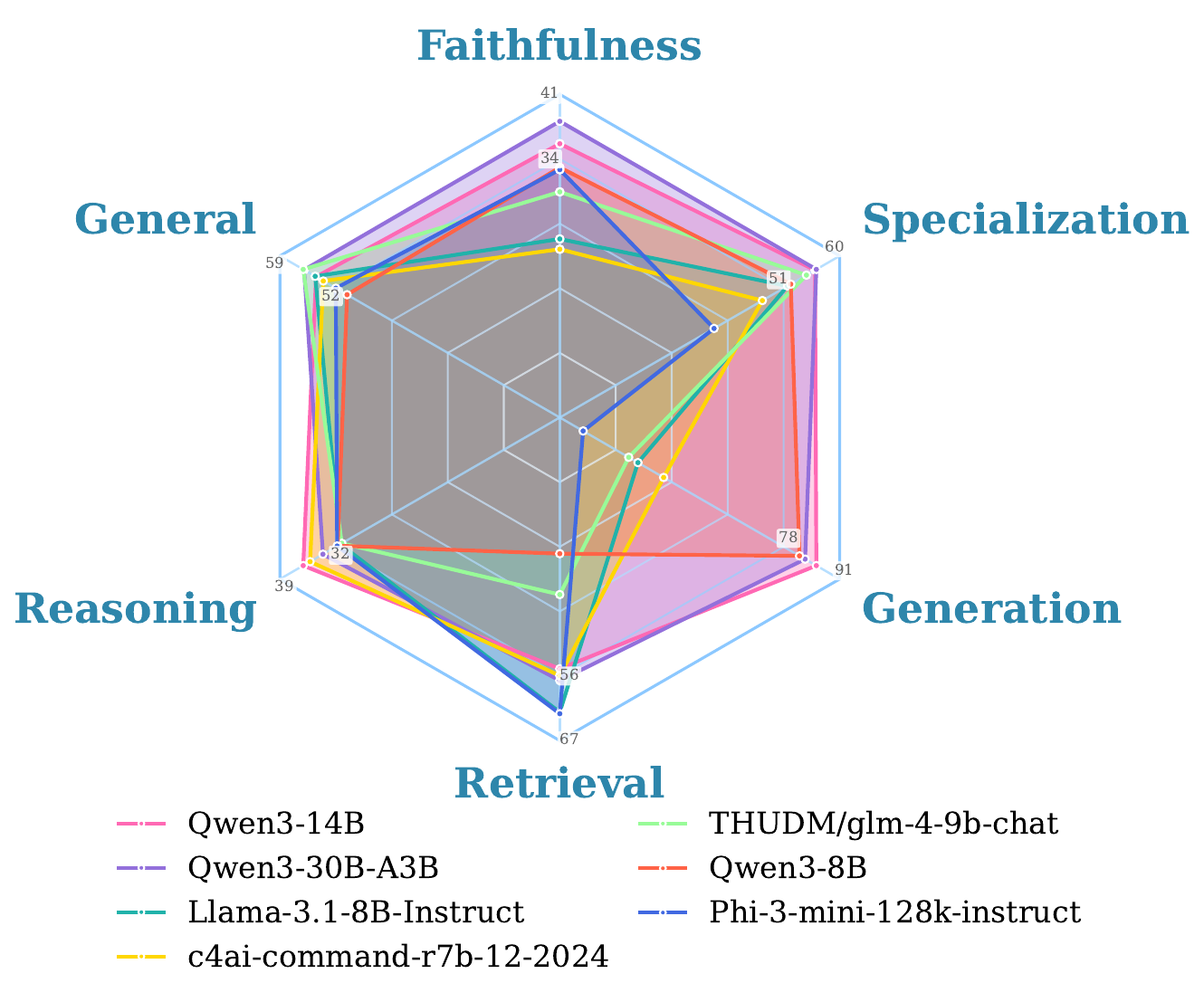}
    \caption{Capability radar chart of partial LCLMs.}
    \label{fig:experiments1}
\end{figure}

\begin{figure}[t]
    \centering
    \includegraphics[width=\linewidth]{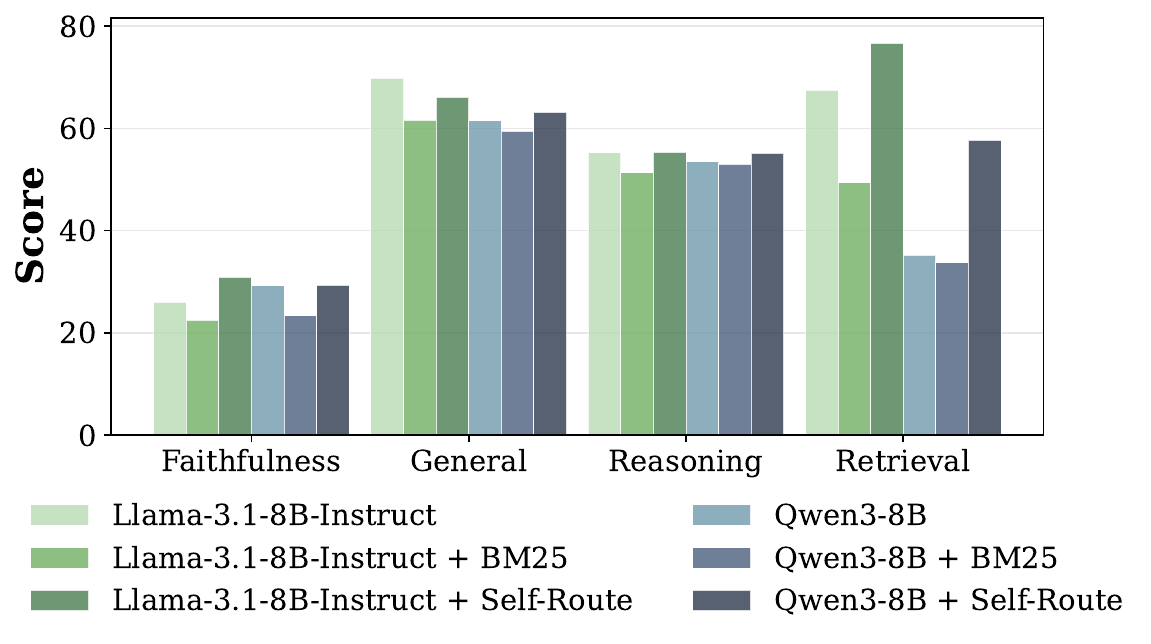}
    \caption{RAG results for partial models. Full evaluation results are shown in Appendix~\ref{appdix:full_exp_res}~(Figure~\ref{fig:rag_eval_total}).}
    \label{fig:rag_eval}
\end{figure}

\subsection{LOOM-Scope Usage}
LOOM-Scope is primarily controlled via user-defined configuration files, enabling a high degree of customization.
As illustrated in Figure~\ref{fig:command_line}, the platform supports evaluation through \textbf{two low-code interfaces}: a command-line interface and a local WebUI, together requiring no more than \textbf{11 lines} of code in total.
For the command-line interface, detailed hyperparameter documentation is available on the official repository\footnote{\url{https://github.com/LCM-Lab/LOOM-Scope}}, allowing users to create custom configuration files to fully control the evaluation workflow.
The WebUI interface, shown in Figure~\ref{fig:screen_shot_webui}, offers an intuitive alternative for users who prefer a graphical interface.

\subsection{Experimental Results}
We evaluate on \textsc{LOOMBench} with three settings: (1) naive LCLM with HF\_Models server, (2) RAG, and (3) inference acceleration methods.
For RAG, we adopt both the BM25 and StreamingRAG algorithms with a retrieval chunk size of 16K and evaluate them on four tasks sampling from \textsc{LOOMBench}.
For inference acceleration methods, experiments are conducted on 128K-length tasks to demonstrate the efficiency.

\paragraph{Naive HF\_Model Results}
We show the partial results in Figure~\ref{fig:experiments1}, where it can be observed that the Qwen-3 series models~\citep{qwen3} exhibit comprehensive long-context capabilities, while other models, e.g., Phi-3~\citep{abdin2024phi}, demonstrate relatively strong understanding abilities but struggle with long-form generation performance.
More evaluation results are shown in Appendix~\ref{appdix:full_exp_res}.

\begin{table*}[t]
    \centering
    \small
    \resizebox{\textwidth}{!}{\begin{tabular}{l | c c c c c c}
    \toprule
    \bf Framework & \bf \makecell[c]{Benchmark$^{\dagger}$\\Num} & \bf \makecell[c]{Model\\Architecture} & \bf \makecell[c]{Interface\\Type}  &  \bf \makecell[c]{Augmentation\\Method} &\bf Inference Engine & \bf \makecell[c]{Custom\\Benchmark} \\
    \midrule
    OpenCompass~\citep{2023opencompass} & 15 & Transformer & Command & - & VLLM / LMDeploy / API & - \\
    EvalHardness~\citep{eval-harness} & 6 & Transformer / Mamba & Command & - & VLLM / SGLang / API & - \\
    UltraEval~\citep{he2024ultraeval} & 5 & Transformer & Command & - & VLLM / API & - \\
    TAIL~\citep{gu-etal-2024-tail} & 1 & - & Command & - & VLLM / API & TAIL \\
    \midrule
    LOOM-Scope~(Ours) & \colorbox{green!50}{22} & \colorbox{green!50}{\makecell[c]{Transformer /\\RNN-series /\\ Linear Attn}} & \makecell[c]{Command /\\WebUI} & \colorbox{green!50}{\makecell[c]{Sparse Attn /\\KV Cache / \\ \textit{etc.}}} & VLLM / SGLang / API  & \colorbox{green!50}{\textsc{LOOMBench}} \\
    \bottomrule
    \end{tabular}}
    \caption{Comparison with other frameworks. $^{\dagger}$ denotes that we only count long-context benchmarks.}
    \label{tab:comparison_other_framework}
\end{table*}

\begin{figure}[t]
  \includegraphics[width=\columnwidth]{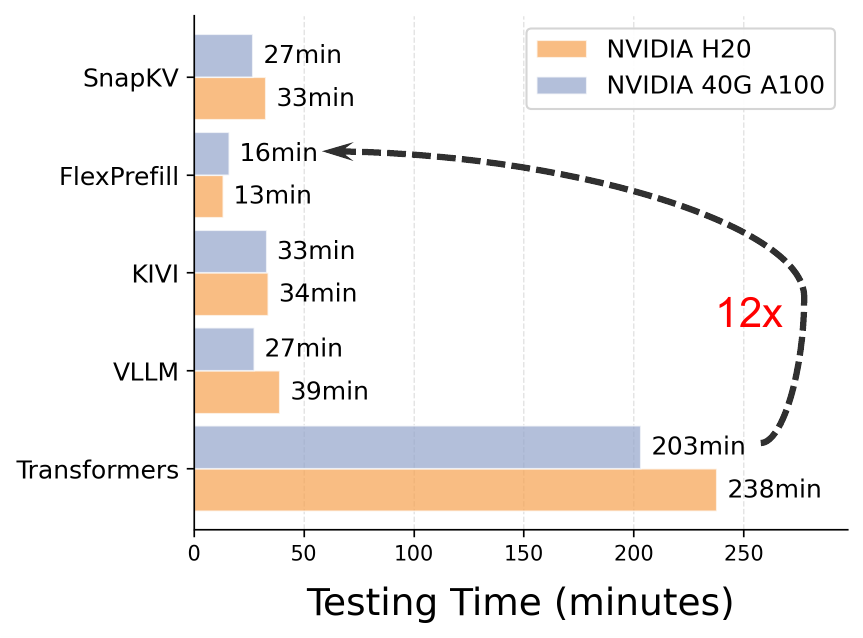}
  \caption{Time cost of acceleration methods.}
  \vspace{-1em}
  \label{fig:prediction_time_acceleration}
\end{figure}

\paragraph{RAG Results}
We experiment with RAG on two settings: rule-based method~(BM25) and model-based method~(Self-Route~\citep{li2024retrieval}).
As shown in Figure~\ref{fig:rag_eval}, we observe that the rule-based RAG method underperforms compared to directly using LCLMs.
In contrast, prediction based on the model-based method~(Self-Route) can improve the performance.
More RAG evaluation results are shown in Appendix~\ref{appdix:full_exp_res}~(Figure~\ref{fig:rag_eval_total}).

\paragraph{Acceleration Method Results}
We optimized and processed selected acceleration methods, ensuring that all can process the 128K-length context on a single 40GB A100 GPU with Llama-3.1-8B-instruct.
After optimization, we evaluated RULER using the LOOM-Scope framework, sampling 15 data instances per subtask under each method's official configuration: the Native Transformer (FlashAttention implementation) used batch size 1, while all acceleration methods used batch size 8.
The timing results for the methods tested on 40GB A100 and H20 GPUs are shown in Figure~\ref{fig:prediction_time_acceleration}. 
The remaining results are shown in Appendix~\ref{appdix:full_exp_res}~(Figure~\ref{fig:reasoning_time_comparison_all}).

\subsection{Comparison with Existing Frameworks}
We compare LOOM-Scope with other evaluation frameworks in Table~\ref{tab:comparison_other_framework}.
We find that LOOM-Scope offers the most comprehensive support for long-context evaluation in terms of benchmark coverage, model architecture compatibility, server deployment, and an integrated comprehensive evaluation benchmark.
Notably, LOOM-Scope is the only existing platform that incorporates long-context inference acceleration methods, greatly improving its extensibility and evaluation efficiency.

\section{Conclusion}
\label{sec:conclusion}
To ensure fairness and efficiency in long-context evaluation, we propose LOOM-Scope, a comprehensive and efficient evaluation framework for long-context models, context lengths ranging from 8K to 2M tokens. 
It supports all mainstream model architectures, provides two user-friendly interfaces (command-line and WebUI), and accommodates various augmentation strategies, including both inference acceleration and retrieval-augmented generation (RAG) methods. 
Experimental results using LOOM-Scope and \textsc{LOOMBench} demonstrate the efficiency and comprehensiveness of our platform.
\section*{Limitation and Future Work}
\label{sec:limitation}
We list two main limitations and their corresponding future work for LOOM-Scope.
\paragraph{Supporting More Benchmarks}
Although our evaluation framework is designed to be modular and extensible, it currently does not include all existing public benchmarks due to the need for customized data formatting and integration. 
Incorporating a broader range of benchmarks to support more flexible and diverse evaluation is one of our key future directions.
\paragraph{Supporting More Modalities}
In addition, the current version of our framework is limited to text evaluation. In future work, we plan to extend support to more modalities relevant to long-context scenarios, such as video~\citep{tang2025video}, along with corresponding inference acceleration techniques. 
These enhancements aim to make our framework more comprehensive and widely applicable across multi-modal long-context tasks.

\section*{Ethics Statement and System License}
\label{sec:ethics_statement}
The development and deployment of LOOM-Scope are guided by ethical principles, with a strong commitment to transparency, reproducibility, and accessibility.
We aim to ensure that the framework is used responsibly and contributes positively to the broader research community. To this end, \textbf{we publicly release the source code and datasets under the \textcolor{blue}{Apache License 2.0}\footnote{\url{https://www.apache.org/licenses/LICENSE-2.0}}
}, which permits open-source use, modification, and commercial deployment, including usage within industrial and enterprise settings.

\section*{Broader Impact Statement}
\label{sec:board_impact}

\paragraph{Transparency and Accountability} All datasets used in LOOM-Scope are clearly labeled with their sources and are licensed under permissive open-source agreements, e.g., Apache 2.0, allowing for modification and redistribution. Our benchmark suite, \textsc{LOOMBench}, builds on these datasets and is fully documented and reproducible. We also provide testing scripts and demo inputs to ensure that all reported results can be independently validated, fostering trust and transparency in long-context evaluation.

\paragraph{Support for Low-Resource Settings} LOOM-Scope is the first framework to support multiple inference acceleration methods under a unified environment, making it particularly suitable for low-resource settings. Users can efficiently evaluate long-context performance and compare trade-offs across various acceleration techniques without the need for large-scale infrastructure.

\paragraph{User Accessibility} To broaden accessibility, LOOM-Scope provides a locally deployable WebUI, enabling researchers and practitioners, regardless of their programming proficiency, to conduct long-context evaluations with minimal setup. This user-centered design lowers the barrier to entry for non-expert users and facilitates practical adoption.

\paragraph{Contribution to the Long-Context Community} We believe that LOOM-Scope will significantly advance research in long-context modeling, an increasingly important direction in the development of large language models~(LLMs). By offering a unified, extensible, and reproducible evaluation platform, we hope to support the community in building more capable, robust, and transparent LCLMs.

\bibliography{main}

\appendix
\clearpage

\section{Related work}
\label{section:related_work}

\subsection{Long-context Augmentation Methods}
To alleviate computational inefficiencies of transformer-based models during long-context inference, recent advances have introduced various augmentation methods, such as RAG~\citep{lee2024longcontextlanguagemodelssubsume,gao2023retrieval,Liu_LlamaIndex_2022}, key-value cache optimization~\citep{li2024survey}, and inference-time sparse attention~\citep{jiang2024minference,laiflexprefill,xu2025xattention,zhang2025spargeattn}.
Among these methods, key-value cache optimization has emerged as the most commonly adopted approach, which can be further categorized into selective eviction~\citep{H2O,Ge2023ModelTY,SnapKV,CAKE}, quantization~\citep{KIVI,KVQuant,OTT,TailorKV}, and approximation~\citep{GEAR,KVMerger,LESS} methods. Notably, LOOM-Scope already supports most mainstream long-context augmentation methods currently in use.

\subsection{Long-context Evaluation}
Existing long-context benchmarks can generally be classified into 3 categories according to the specific capabilities they assess: long-context understanding~\citep{bai2023longbench,zhang2024bench,robertsneedle}, long-form reasoning~\citep{kuratov2024babilong,bai2024longbench,song2024counting}, and long-form generation~\citep{bai2024longwriter}.
Furthermore, the task formats themselves can be further divided into real-world tasks~(e.g., multi-document QA~\citep{wang2024leave,yuan2024lv}, article generation~\citep{liu2024longgenbench}) and synthetic tasks (e.g., needle-in-a-haystack retrieval~\citep{yu2025sequential,wang2024needle}, synthetic multi-hop reasoning~\citep{yu2024hyper}).
Currently, there are more than 150 benchmarks dedicated to long-context evaluation~\citep{liu2025comprehensive}. 
However, conflicting evaluation results exist across benchmarks~\citep{yen2024helmet}, and disparate evaluation settings, such as model prompts and execution environments, make those benchmarks cumbersome to use and difficult to compare. 
In this paper, we introduce LOOM-Scope, a comprehensive and efficient framework for long-context evaluation, offering a unified environment that enables plug-and-play benchmarking and supports highly user-customizable modules for flexible and convenient evaluation.




\begin{table*}[htbp]
\centering
\renewcommand{\arraystretch}{1.5}
\resizebox{\linewidth}{!}{ 
\begin{tabular}{ccll}
\toprule
\multicolumn{2}{c}{\textbf{Type}} & \textbf{Implementation} & \textbf{Description} \\
\midrule
\multirow{4}{*}{\rotatebox{90}{\textbf{Server}}} 
    & \multirow{3}{*}{\textbf{Local Models}} 
        & HF\_Models & HuggingFace optimized inference \\
        \cline{3-4}
    & & VLLM & PagedAttention for long sequences \\
        \cline{3-4}
    & & SGLang & RadixAttention for fast execution \\
    \cline{2-4}
    & \textbf{Cloud Models} 
        & API & API-based scalable inference service \\
\midrule
\multirow{7}{*}{\rotatebox{90}{\textbf{Architecture}}} 
    & \multirow{2}{*}{\textbf{RNN-based}} 
        & RWKV & Attention-free recurrent architecture \\
        \cline{3-4}
    & & Mamba & Selective state space model \\
    \cline{2-4}
    & \textbf{Linear-Attention} 
        & GLA & Gated linear attention \\
    \cline{2-4}
    & \multirow{4}{*}{\textbf{Transformer-based}} 
        & QWEN & Pretrained LLM proposed by Alibaba Cloud \\
        \cline{3-4}
    & & Deepseek & Pretrained LLM proposed by Deepseek \\
        \cline{3-4}
    & & Mistral & Frontier AI LLMs, Assistants, Agents, Services \\
        \cline{3-4}
    & & GLM & Open Multilingual Multimodal Chat LMs \\
\midrule
\multirow{14}{*}{\rotatebox{90}{\textbf{Augmentation}}} 
    & \multirow{4}{*}{\textbf{Token Eviction}} 
        & H2O & Attention-based selection \\
        \cline{3-4}
    & & StreamingLLM & Retain first few tokens \\
        \cline{3-4}
    & & SnapKV & Attention Pooling before selection \\
        \cline{3-4}
    & & L2Compress & L2 Norm is better than attention as a metric \\
    \cline{2-4}
    & \multirow{2}{*}{\textbf{Layer-wise}} 
        & PyramidKV & Layer-wise budget allocation \\
        \cline{3-4}
    & & CakeKV & Layer-specific preference score \\
    \cline{2-4}
    & \textbf{Quantization} 
        & KIVI & Asymmetric 2-bit Quantization \\
    \cline{2-4}
    & \textbf{Hybrid} 
        & ThinK & Thinner Key Cache by Query-Driven Pruning \\
    \cline{2-4}
    & \textbf{Token Merge} 
        & CaM & Cache Merging for LLMs Inference \\
    \cline{2-4}
    & \multirow{2}{*}{\textbf{Sparse Attention}} 
        & FlexPrefill & A context-aware sparse attention mechanism \\
        \cline{3-4}
    & & XAttention & Block sparse attention with antidiagonal scoring \\
    \cline{2-4}
    & \multirow{3}{*}{{\textbf{RAG}}} 
        & BM25 & Classic sparse retrieval algorithm \\
        \cline{3-4}
        && LlamaIndex & Framework for efficient RAG indexing \\
        \cline{3-4}
        && OpenAI & Powers generation/embeddings in RAG \\
    
\bottomrule
\end{tabular}
}
\caption{Deployment Module Details}
\label{tab:deployment_module}
\end{table*}


\section{Benchmark Overview}
\label{appdix:bench_overview}
We provides a systematic overview of the supported benchmarks, categorizing them by capability and summarizing their key metrics and data volumes in Table~\ref{tab:benchmarks}.
\begin{table*}[htbp]
\centering
\resizebox{\linewidth}{!}{ 
\begin{tabular}{ccccc}
\toprule
\textbf{Category} & \textbf{Benchmark} & \textbf{Metric} & \textbf{Data Volumn}& \textbf{Data Length}\\
\midrule

\multirow[m]{2}{*}{Faithfulness} 
 & L-CiteEval & ROUGE,ACC,Recall& 3220& 0 to 52K\\
 & LongCite & F1,Recall,Precision,ACC,LLM & 1000 & 0 to 82K \\

\midrule
\multirow[m]{5}{*}{General}
 & LEval & ACC,LLM,ROUGE,F1& 8645& 3K to 200K\\
 & LooGLE & ROUGE,BLEU,LLM& 1617& 12K to 282K\\
 & RULER & ACC,F1,Recall& -& 4K to 128K\\
 & LongBench & ACC,F1& 8418& 0 to 64K\\
 & BAMBOO& ACC,F1,Recall,Pass@1& 2904& 0 to 16K\\

\midrule
\multirow[m]{6}{*}{Reasoning}
 & Counting-Stars & F1,Pass@N& 128 & 4K to 128K\\
 & LongIns & F1& 20767& 0 to 16K\\
 & LVEval & ROUGE,F1& 8645& 8K to 418K\\
 & LongBench V2 & ACC& 503& 8K to 4M\\
 & babilong & ACC& 4500& 0 to 128K\\
 & Ada\_LEval & ACC& 13200& 0 to 141K\\

\midrule
\multirow[m]{4}{*}{Retrieval}
 & NIAH & ACC& -& - \\
 & NThread & ACC,LLM& 10860& 1K to 616K\\
 & InfiniteBench & ROUGE,F1& 3946 &26k to 5M\\
 & NoLiMa & ACC& -& 1K to 32K\\

\midrule
\multirow[m]{1}{*}{Generation}
 & LongWriter & LLM& 228& 0 to 1K\\

\midrule
\multirow[m]{4}{*}{Specialization}
 & LIBRA & F1,EM& 13071& 1K to 142K\\
 & CLongEval & F1,ROUGE,ACC& 7263& 1K to 128K\\
 & LongHealth & ACC& 6000& 8K to 16K\\
 & LongSafety & F1,LLM& 6172& 3K to 22K\\
\bottomrule
\end{tabular}
}
\caption{Benchmarks Overview.The hyphen ("-") represents infinite length.}
\label{tab:benchmarks}
\end{table*}

\subsection{Supporting Benchmarks}
 \begin{table*}[htbp]
\centering
\begin{center}
\resizebox{\linewidth}{!}{ 
\begin{tikzpicture}
\node[draw=black!70, fill=gray!10, rounded corners, inner sep=8pt] {
\begin{tabular}{@{}l@{\hspace{15pt}}l@{\hspace{15pt}}l@{\hspace{15pt}}l@{}}
\textbf{Total Benchmarks:} 22 & \textbf{Total Tasks:} 149 & \textbf{Languages:} EN, ZH, RU & \textbf{Context Length:} 8K-2M tokens
\end{tabular}
};
\end{tikzpicture}
}
\end{center}

\renewcommand{\arraystretch}{1.5}
\resizebox{\linewidth}{!}{ 
\begin{tabular}{l|l|c|c|c}
\toprule
\textbf{Category} & \textbf{Benchmark} & \textbf{Tasks} & \textbf{Key Capabilities} & \textbf{Domains} \\
\midrule

\multirow{2}{*}{Faith.} 
& \highlight{L\_CiteEval} & 11 & Citation evaluation, multi-hop QA, dialogue simulation & News, Government, Dialogue \\
& LongCite & 1 & Long document citation accuracy & Academic \\
\midrule

\multirow{5}{*}{General}
& BAMBOO & 10 & Paper QA, hallucination detection, meeting/show prediction & Academic, Business \\
& \highlight{LongBench} & 16 & Multilingual QA, summarization, passage retrieval, code & Multi-domain \\
& \highlight{LEval} & 20 & Financial QA, legal contracts, scientific papers, TV shows & Finance, Legal, Science \\
& \highlight{RULER} & 5 & Variable tracking, multi-query NIAH & Synthetic \\
& LooGLE & 2 & Long dependency QA, summarization & General \\
\midrule

\multirow{4}{*}{Retrieve}
& \highlight{NIAH} & 1 & Classic needle-in-haystack retrieval & Synthetic \\
& NThread & 6 & Multi-needle variants with CoT and distractors & Synthetic \\
& \highlight{InfiniteBench} & 12 & Code debug, math calc, long dialogue, book QA (EN/ZH) & Code, Math, Literature \\
& NoLiMa & 5 & Conditional needles, multi-thread tracking & Synthetic \\
\midrule

Generation & \highlight{LongWriter} & 3 & Long-form writing in English and multilingual & Creative Writing \\
\midrule

\multirow{6}{*}{Reasoning}
& \highlight{Counting-Stars} & 4 & Reasoning and searching in EN/ZH & Logic \\
& \highlight{babilong} & 5 & Multi-hop reasoning tasks (qa1-qa5) & Synthetic \\
& LongIns & 2 & GIST extraction, LIST processing & Instruction Following \\
& \highlight{LVEval} & 11 & Mixed-up tasks, fact recall, reading comprehension & Multi-domain \\
& \highlight{LongBench\_v2} & 1 & Deep understanding with 8K-2M context & Complex Reasoning \\
& Ada-LEval & 2 & Stack selection, text sorting & Algorithmic \\
\midrule

\multirow{4}{*}{Special.}
& \highlight{LIBRA} & 21 & Russian language tasks across all categories & Russian Multi-domain \\
& CLongEval & 7 & Memory, summarization, table query, key retrieval & Chinese-focused \\
& LongHealth & 3 & Medical document understanding & Healthcare \\
& LongSafety & 1 & Safety evaluation in long contexts & AI Safety \\
\bottomrule
\end{tabular}
}
\caption{Comprehensive Overview of Long Context Evaluation Benchmarks.\textsc{LOOMBench} is indicated by the orange-highlighted entries}
\label{tab:benchmark_details}
\end{table*}
 LOOM-Scope supports 22 widely used long-context benchmarks in Table~\ref{tab:benchmark_details}, covering more than 140 subtasks, the 8k-2M context range, and spanning six main LCLM capabilities:
\begin{itemize}
\setlength\itemsep{0em}
\item \textbf{Faithfulness}: Focus on citation accuracy and information attribution
\item \textbf{General}: Broad evaluation across multiple domains and task types
\item \textbf{Retrieve}: Needle-in-haystack and information location tasks
\item \textbf{Generation}: Long-form content creation capabilities
\item \textbf{Reasoning}: Complex multi-hop reasoning and logical inference
\item \textbf{Specialization}: Domain-specific (medical, legal) and language-specific benchmarks
\end{itemize}

\subsection{Statistic of \textsc{LOOMBench}}
\label{appdix:statistic_loombench}
\begin{figure}[t]
  \includegraphics[width=\columnwidth]{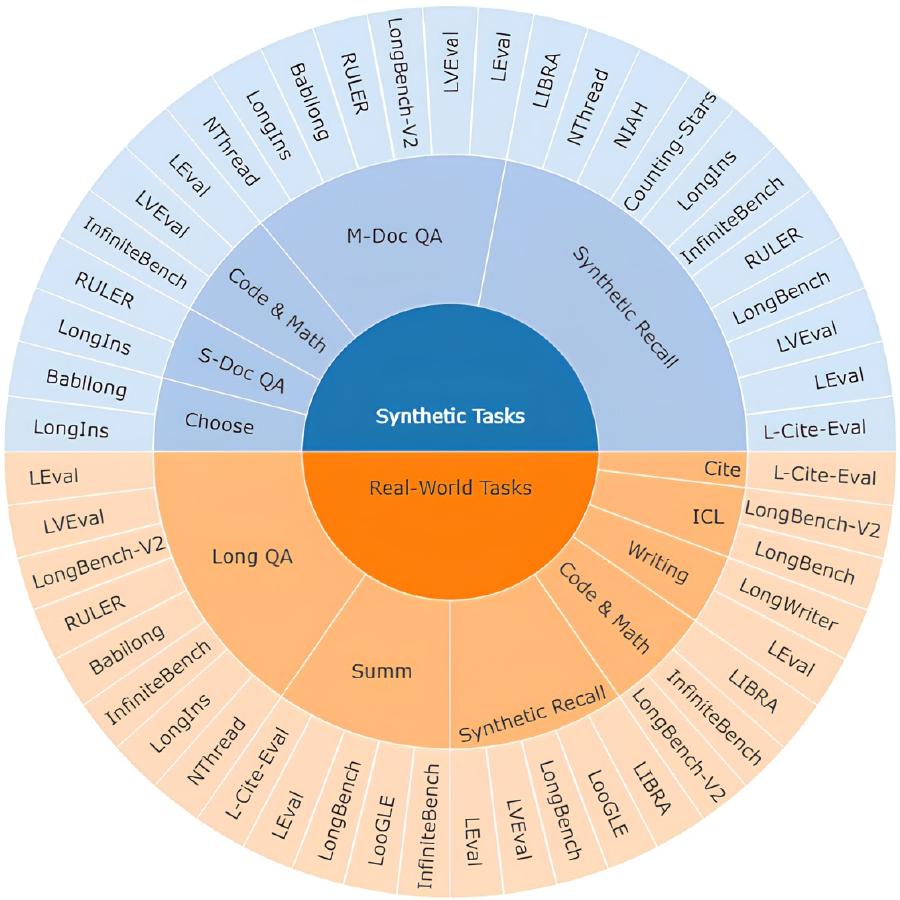}
  \caption{Task distribution of \textsc{LOOMBench}.}
  \label{fig:loombench}
\end{figure}

\textsc{LOOMBench} is a composite benchmark constructed from 12 existing datasets, as indicated by the orange-highlighted entries in Table~\ref{tab:benchmark_details}. These selected datasets ensure a balanced evaluation across six core long-context capabilities, providing comprehensive coverage of the benchmark's design objectives.
The visualization of task distribution is also shown in Figure~\ref{fig:loombench}.

\section{Full Experimental Results}
\label{appdix:full_exp_res}
Table~\ref{tab:loombench_results} presents the full evaluation of all models in \textsc{LOOMBench}, including detailed performance metrics for LCLMS. Table~\ref{tab:acceleration_performance} provides comparative results of the augmentation method. These results collectively demonstrate the benchmark's comprehensive assessment of long-context capabilities.

\subsection{Vanilla Model Performance}
\begin{figure}[h]
  \includegraphics[width=\columnwidth]{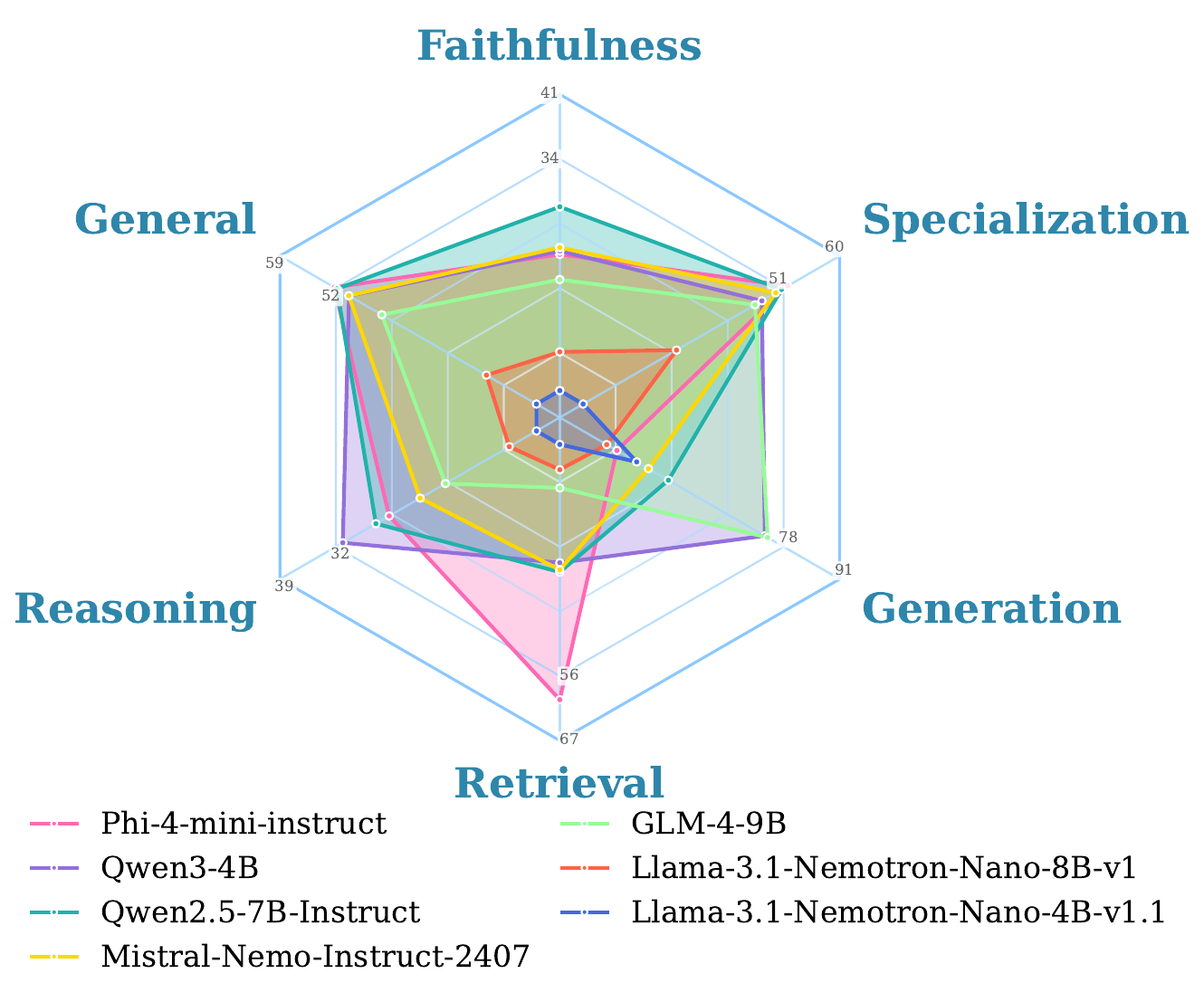}
  \caption { Capability radar chart of remain LCLMs. }
  \label{fig:experiments2}
\end{figure}

\paragraph{Model Performance} We evaluated the six core capabilities of 14 mainstream LCLMs in \textsc{LOOMBench} using the LOOM-Scope framework, with results presented in Figure~\ref{fig:experiments1}, Figure~\ref{fig:experiments2}, and Table~\ref{tab:loombench_results}.  

\begin{table*}[htbp]
\centering
\resizebox{\linewidth}{!}{ 
\begin{tabular}{c|cc|c|ccc|cccc|cc|c|c}
\toprule
\multirow{2}{*}{\textbf{Rank}} & \multirow{2}{*}{\textbf{Model}} & \multirow{2}{*}{\textbf{Avg}} & \textbf{Faithfulness} & \multicolumn{3}{c}{\textbf{General}} & \multicolumn{4}{c}{\textbf{Reasoning}} & \multicolumn{2}{c}{\textbf{Retrieve}} & \textbf{Generation}& \textbf{Specialization}\\

 & & &\textbf{LCite} & \textbf{LEval} & \textbf{RULER} & \textbf{LongB} & \textbf{BABI} & \textbf{Count} & \textbf{LVE} & \textbf{LB2} & \textbf{NIAH} & \textbf{InfB} & \textbf{LongW} & \textbf{LIBRA}\\
\midrule
\multicolumn{15}{c}{\textit{Qwen Series}} \\
1 & Qwen3-14B & 51.54 & 35.64 & 43.84 & 74.94 & 45.47 & 59.15 & 56.41 & 21.26 & 29.85 & 100 & 10.24 & 85.75 & 55.87 \\
2 & Qwen3-30B-A3B & 51.18 & 37.96 & 40.61 & 78.32 & 43.24 & 60.31 & 48.96 & 22.82 & 28.42 & 100 & 14.14 & 83.24 & 56.09 \\
6 & Qwen3-8B & 44.71 & 33.18 & 41.15 & 67.68 & 38.62 & 55.28 & 52.32 & 15.15 & 27.25 & 64.00 & 8.06 & 81.99 & 51.78 \\
9 & Qwen3-4B & 43.10 & 24.55 & 39.03 & 70.29 & 39.32 & 55.01 & 42.06 & 18.24 & 32.52 & 62.00 & 13.05 & 74.25 & 46.92 \\
10 & Qwen2.5-7B-Instruct & 42.01 & 29.12 & 44.63 & 72.02 & 40.85 & 55.89 & 38.25 & 14.94 & 27.33 & 64.18 & 13.97 & 52.75 & 50.23 \\
\midrule
\multicolumn{15}{c}{\textit{Meta/Llama Series}} \\
3 & Llama-3.1-8B-Instruct & 46.94 & 25.79 & 39.70 & 86.79 & 37.94 & 57.42 & 37.68 & 25.66 & 30.40 & 91.00 & 33.64 & 45.96 & 51.24 \\
13 & Nemotron-Nano-8B-v1 & 24.47 & 14.11 & 34.32 & 42.51 & 27.19 & 28.78 & 11.72 & 6.57 & 12.67 & 43.73 & 0.47 & 38.99 & 32.54 \\
14 & Nemotron-Nano-4B-v1.1 & 21.05 & 10.11 & 25.88 & 38.85 & 19.94 & 22.67 & 7.48 & 6.69 & 22.67 & 28.38 & 7.43 & 45.68 & 16.81 \\
\midrule
\multicolumn{15}{c}{\textit{Microsoft Phi Series}} \\
7 & Phi-3-mini-128k-instruct & 44.67 & 32.96 & 39.87 & 78.62 & 38.31 & 53.56 & 31.04 & 39.87 & 24.02 & 90.00 & 35.14 & 33.73 & 38.86\\
8 & Phi-4-mini-instruct & 43.83 & 24.20 & 40.18 & 76.70 & 42.69 & 53.56 & 13.31 & 30.93 & 31.33 & 92.61 & 27.87 & 41.27 & 51.28 \\
\midrule 
\multicolumn{15}{c}{\textit{GLM Series}} \\
5 & GLM-4-9B-chat & 44.89 & 30.66 & 46.42 & 85.25 & 45.24 & 55.00 & 36.84 & 23.33 & 32.00 & 65.27 & 20.35 & 43.90 & 54.42 \\
12 & GLM-4-9B & 36.80 & 21.59 & 45.70 & 55.96 & 38.41 & 46.33 & 21.51 & 17.18 & 24.00 & 47.15 & 3.11 & 74.89 & 45.76 \\
\midrule 
\multicolumn{15}{c}{\textit{Other Models}} \\
4 & c4ai-command-r7b-12-2024 & 45.39 & 24.73 & 42.68 & 77.41 & 37.16 & 47.44 & 35.00 & 35.66 & 33.33 & 92.43 & 20.09 & 51.69 & 47.00 \\
11 & Mistral-Nemo-Instruct-2407 & 38.37 & 24.91 & 42.47 & 60.60 & 39.75 & 53.67 & 21.12 & 21.61 & 21.34 & 60.41 & 16.98 & 48.30 & 49.25 \\

\bottomrule
\end{tabular}
}
\caption{High-performance Long-context Language Models on LOOMBench: Comprehensive evaluation across 12 benchmarks measuring reasoning, retrieval, generation, and comprehension capabilities.
\textbf{LCite}: L\_CiteEval; \textbf{LongB}: LongBench; \textbf{BABI}: BABILong; \textbf{Count}: Counting-Stars; \textbf{LVE}: LVEval; \textbf{LB2}: LongBench\_v2; \textbf{InfB}: InfiniteBench; \textbf{LongW}: LongWriter}
\label{tab:loombench_results}
\end{table*}

\paragraph{Evaluation Latency}
We used the LOOM-Scope to evaluate both the native benchmark and \textsc{LOOMBench}. The Evaluation Latency presented in Table~\ref{tab:bench_time} clearly demonstrates that \textsc{LOOMBench}, which serves as a comprehensive but lightweight benchmarking tool, enables efficient performance assessment.
\begin{table*}[htbp]
\centering
\resizebox{\linewidth}{!}{ 
\begin{tabular}{c|c|c|ccc|cccc|cc|c|c}
\toprule
\multirow{2}{*}{\textbf{Benchmark}} & \multirow{2}{*}{\textbf{Sum}} & \textbf{Faithfulness} & \multicolumn{3}{c}{\textbf{General}} & \multicolumn{4}{c}{\textbf{Reasoning}} & \multicolumn{2}{c}{\textbf{Retrieve}} & \textbf{Generation}& \textbf{Specialization}\\

 & &\textbf{LCite} & \textbf{LEval} & \textbf{RULER} & \textbf{LongB} & \textbf{BABI} & \textbf{Count} & \textbf{LVE} & \textbf{LB2} & \textbf{NIAH} & \textbf{InfB} & \textbf{LongW} & \textbf{LIBRA}\\
\midrule
\multicolumn{14}{c}{\textit{NVIDIA GeForce RTX 3090}} \\
\textsc{LOOMBench} & 22:42:34 & 2:20:39 & 2:17:58 & 1:59:01 & 1:40:47 & 2:19:20 & 1:07:38 & 2:35:32 & 1:19:12 & 0:50:40 & 1:49:10 & 1:47:06 & 2:35:31 \\
Native & 216:44:36 & 10:49:25 & 6:09:10 & 35:24:45 & 8:28:28 & 31:8:45 & 0:39:04 & 37:12:43 & 3:25:30 & 4:31:48 & 45:33:55 & 9:04:27 & 1:34:02 \\

\midrule
\multicolumn{14}{c}{\textit{NVIDIA 40GB A100}} \\
\textsc{LOOMBench} & 8:38:06 & 0:25:53 & 1:10:46 & 0:59:17 & 0:14:57 &0:41:49&0:37:15&0:50:53&0:48:50&0:24:28&0:39:12&1:06:59 & 0:22:07\\
Native & 110:56:06 & 7:35:33 & 3:58:46 & 16:40:31 & 4:00:00 & 12:09:37 & 0:19:38 & 21:01:12 & 1:37:19 & 1:59:01 & 23:40:31 & 2:28:56 & 0:43:07 \\

\midrule
\multicolumn{14}{c}{\textit{NVIDIA H20}} \\
\textsc{LOOMBench} & 5:40:40 & 0:29:11 & 0:32:59 & 0:31:57 & 0:14:19 & 0:30:24 & 0:20:49 & 0:31:49 & 0:26:01 & 0:34:08 & 0:24:24 & 0:33:15 & 0:31:24 \\
Native & 96:14:11 & 6:45:38 & 3:41:35 & 20:10:37 & 6:25:36 & 16:37:10 & 0:17:58 & 23:42:26 & 2:05:23 & 2:50:34 & 26:01:28 & 1:29:10 & 0:48:51 \\

\bottomrule
\end{tabular}
}
\caption{The Evaluation Latency for the Native Benchmark and \textsc{LOOMBench}.In \textsc{LOOMBench}, the naive Transformer service running on a single 40GB A100 GPU encountered Out-of-Memory (OOM) errors on certain datasets. To address this, we employed a dual-GPU inference strategy for those datasets, which resulted in slightly longer processing times.
\textbf{LCite}: L\_CiteEval; \textbf{LongB}: LongBench; \textbf{BABI}: BABILong; \textbf{Count}: Counting-Stars; \textbf{LVE}: LVEval; \textbf{LB2}: LongBench\_v2; \textbf{InfB}: InfiniteBench; \textbf{LongW}: LongWriter}
\label{tab:bench_time}
\end{table*}

\subsection{Augmentation Methods}
\begin{figure*}[h]
  \includegraphics[width=\linewidth]{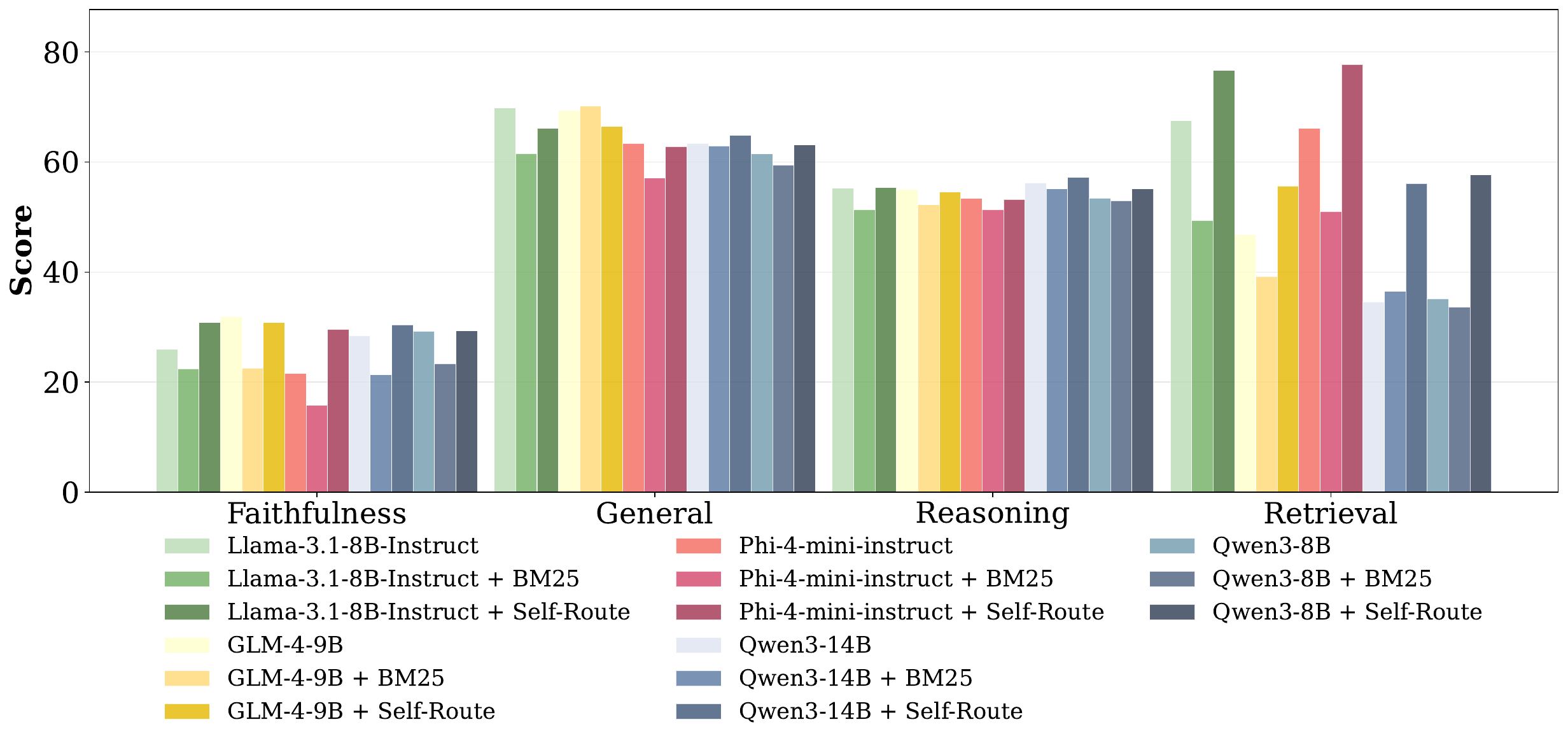}
  \caption {Complete RAG test results for mainstream models.}
  \label{fig:rag_eval_total}
\end{figure*}

\begin{figure*}[h]
  \includegraphics[width=\linewidth]{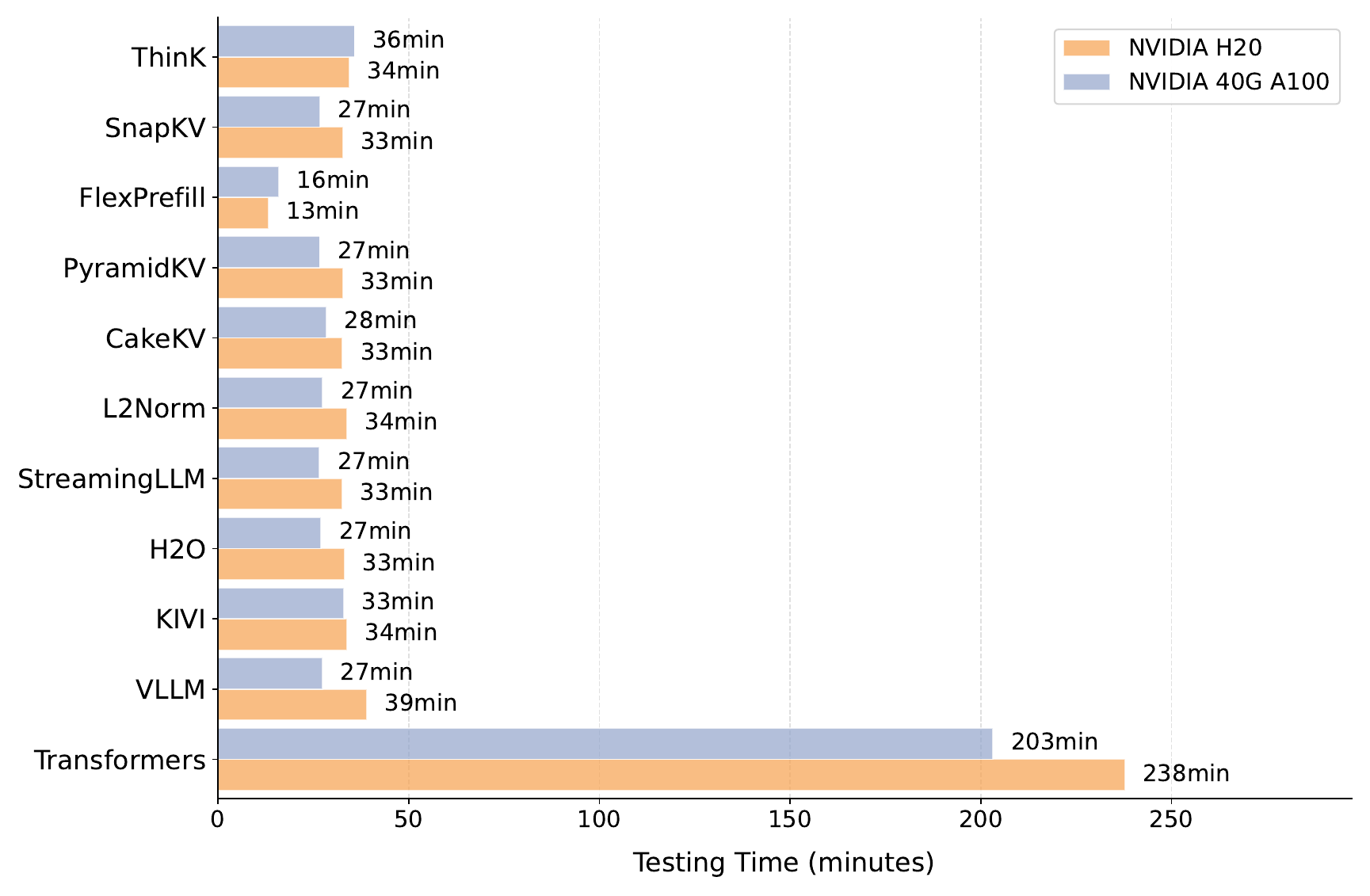}
  \caption {Complete augmentation methods test results for Llama-3.1-8B-Instruct models.}
  \label{fig:reasoning_time_comparison_all}
\end{figure*}

To systematically investigate the efficacy of long-context augmentation techniques, we evaluated retrieval-augmented generation and inference acceleration methods across mainstream methods using LOOM-Scope. Figure~\ref{fig:rag_eval_total} presents comprehensive comparisons of RAG performance. Figure~\ref{fig:reasoning_time_comparison_all} further quantifies the latency reductions for Llama-3.1-8B-Instruct under various acceleration strategies, demonstrating up to \textbf{12×} speedup with acceleration methods. Table~\ref{tab:acceleration_performance} and Table~\ref{tab:xattn_performance} consolidate a detailed result of the Performance and Prediction Time. 

\begin{table*}[htbp]
\centering
\setlength{\tabcolsep}{4pt}
\renewcommand{\arraystretch}{1.5}
\resizebox{\linewidth}{!}{ 
\begin{tabular}{lc|ccccccccccccc}
\toprule
\multicolumn{2}{c}{\textbf{RULER}} & \textbf{KIVI} & \textbf{H2O} & \textbf{StreamingLLM} & \textbf{L2Norm} & \textbf{CaM} & \textbf{CakeKV} & \textbf{PyramidKV} & \textbf{FlexPrefill} & \textbf{SnapKV} & \textbf{ThinK} & \textbf{Transformers} & \textbf{VLLM} & \textbf{SGLang}\\
\midrule
\multirow{6}{*}{\rotatebox{90}{\textbf{NVIDIA RTX 3090}}} 
 & \textbf{4k}& \orangecell{94.83} & 46.13 & 27.79 & 36.42 & 67.39 & 91.88 & 77.42 & 93.33 & 81.54 & 81.83 & \greencell{94.83} & \orangecell{89.54} & 91.09 \\ 
 & \textbf{8k}& \orangecell{94.29} & 34.42 & 22.21 & 27.75 & 59.60 & 82.42 & 71.25 & 87.92 & 73.29 & 72.88 & \greencell{94.42} & \orangecell{90.74} & 89.83 \\ 
 & \textbf{16k}& \orangecell{91.08} & 21.63 & 11.46 & 27.58 & 48.53 & 78.96 & 68.75 & 83.38 & 67.88 & 67.25 & \greencell{92.50} & \orangecell{84.09} & 86.40 \\ 
 & \textbf{32k}& \orangecell{89.50} & 19.67 & 10.79 & 22.17 & 43.42 & 79.54 & 64.33 & 83.38 & 66.29 & 66.21 & \greencell{92.75} & \orangecell{79.83} & 74.58 \\ 
 & \textbf{64k}& - & 10.54 & 7.67 & 18.25 & 33.25 & 62.88 & 57.29 & 65.04 & 55.58 & 56.92 & \greencell{83.71} & \orangecell{82.32} & 80.98\\ 
 & \textbf{128k}& - & 6.38 & 9.79 & 12.17 & 28.10 & 53.17 & 48.29 & 46.58 & 47.25 & 48.71 & - & \orangecell{36.55} & 67.07 \\ \midrule
 \multicolumn{2}{c}{\textbf{Prediction Time}} &\greencell{1:29:58} & 5:11:21 & 5:07:27 & 5:08:31 & 50:31:13 & 5:13:01 & 5:14:12 & 3:58:17 & 5:08:31 & 5:18:08 & \orangecell{3:11:15} & 1:49:41 & 1:41:35 \\ \midrule
 
\multirow{6}{*}{\rotatebox{90}{\textbf{NVIDIA 40GB A100}}} 
 & \textbf{4k}& 95.46 & 46.38 & 28.83 & 33.96 & 67.29 & 89.67 & 77.00 & 93.58 & 79.75 & 79.33 & \greencell{95.04} & \orangecell{90.71} & 90.76 \\
 & \textbf{8k}& \orangecell{94.42} & 31.96 & 21.58 & 26.83 & 59.67 & 82.46 & 70.79 & 89.04 & 72.67 & 71.92 & \greencell{94.46} & \orangecell{91.63} & 92.43 \\ 
 & \textbf{16k} & \orangecell{90.04} & 23.00 & 10.63 & 27.92 & 48.53 & 77.29 & 67.54 & 84.83 & 66.63 & 65.25 & \greencell{92.50} & \orangecell{91.75} & 85.67 \\
 & \textbf{32k} & \orangecell{89.00} & 19.86 & 10.79 & 20.83 & 43.21 & 76.42 & 65.25 & 79.63 & 63.88 & 66.46 & \greencell{92.08} & \orangecell{86.25} & 77.46 \\
 & \textbf{64k} & \orangecell{90.25} & 10.50 & 7.67 & 17.25 & 33.35 & 65.79 & 55.54 & 62.50 & 54.75 & 55.75 & \greencell{83.71} & \orangecell{78.83} & 82.11 \\
 & \textbf{128k} & \orangecell{67.92} & 6.17 & 10.63 & 12.17 & 28.03 & 55.21 & 49.54 & 49.42 & 50.79 & 51.00 & \greencell{75.21} & \orangecell{17.50} & 69.01 \\ \midrule
\multicolumn{2}{c}{\textbf{Prediction Time}} & \greencell{0:32:57} & 0:27:00 & 0:26:35 & 0:27:26 & - & 0:28:26 & 0:26:39 & 0:15:58 & 0:26:44 & 0:35:46 & \orangecell{3:23:10} & \greencell{0:27:20} & 0:20:27 \\ \midrule

\multirow{6}{*}{\rotatebox{90}{\textbf{NVIDIA H20}}} 
 & \textbf{4k} & \orangecell{93.79} & 54.33 & 28.00 & 34.21 & 66.29 & 89.67 & 77.13 & 92.63 & 82.22 & 80.83 & \greencell{94.00} & \orangecell{91.75} & 94.29\\ 
 & \textbf{8k} & \orangecell{94.33} & 41.25 & 21.58 & 26.75 & 59.67 & 82.63 & 71.50 & 87.88 & 70.46 & 72.33 & \greencell{94.46} & \orangecell{88.79} & 88.83\\ 
 & \textbf{16k} & \orangecell{90.71} & 24.17 & 10.63 & 27.00 & 48.33 & 77.21 & 65.33 & 84.54 & 64.57 & 65.92 & \greencell{92.50} & \orangecell{84.58} & 84.88\\ 
 & \textbf{32k} & \orangecell{88.08} & 19.42 & 10.79 & 22.58 & 43.42 & 76.42 & 66.21 & 80.71 & 69.27 & 66.46 & \greencell{91.88} & \orangecell{79.50} & 78.47\\ 
 & \textbf{64k} & \orangecell{80.91} & 12.33 & 6.83 & 16.58 & 34.25 & 65.71 & 58.25 & 62.96 & 57.73 & 56.25 & \greencell{83.71} & \orangecell{77.79} & 81.63 \\ 
 & \textbf{128k} & \orangecell{69.13} & 10.92 & 10.63 & 12.25 & 28.13 & 54.21 & 47.58 & 48.08 & 45.94 & 47.25 & \greencell{75.21} & \orangecell{57.88} &65.92 \\  \midrule
\multicolumn{2}{c}{\textbf{Prediction Time}} &\greencell{0:33:44} &0:33:08 &0:32:35 &00:33:48 &- &0:32:37 &0:32:41 &0:13:10 &0:32:41 &0:34:25 &\orangecell{3:57:44} &\greencell{0:38:56} & 0:23:18\\
\bottomrule
\end{tabular}}
\caption{Performance comparison of various acceleration methods for the Llama-3.1-8B-Instruct model across different hardware configurations and input scales, showcasing reasoning times and accuracy metrics. The batch size per GPU is 8."-" represent Out of Memory.}
\label{tab:acceleration_performance}
\end{table*}
\begin{table*}[htbp]
\centering

\setlength{\tabcolsep}{4pt}
\renewcommand{\arraystretch}{1.5}
\begin{tabular}{lc|c}
\toprule
 \multicolumn{3}{c}{ \textbf{XAttention}} \\
\midrule

\multirow{6}{*}{\rotatebox{90}{\textbf{NVIDIA 40GB A100}}} & \textbf{4k}& 93.45\\
& \textbf{8k}& 91.00 \\ 
& \textbf{16k} & 89.94 \\
& \textbf{32k} & 84.07 \\
& \textbf{64k} & 78.66 \\
& \textbf{128k} & 34.35 \\ \midrule
\multicolumn{2}{c}{\textbf{Prediction Time}} & 0:16:34 \\

\bottomrule
\end{tabular}
\caption{Performance of XAttention acceleration methods for the Llama-3.1-8B-Instruct model across different input scales in RULER, showcasing reasoning times and accuracy metrics.XAttention only optimized the A100.}
\label{tab:xattn_performance}
\end{table*}

\end{document}